\documentclass[10pt,twocolumn,letterpaper]{article}

\usepackage[pagenumbers]{cvpr} 

\usepackage{graphicx}
\usepackage{amsmath}
\usepackage{amssymb}
\usepackage{booktabs}
\usepackage{multirow}
\usepackage{array}

\usepackage[pagebackref,breaklinks,colorlinks]{hyperref}

\usepackage[capitalize]{cleveref}
\crefname{section}{Sec.}{Secs.}
\Crefname{section}{Section}{Sections}
\Crefname{table}{Table}{Tables}
\crefname{table}{Tab.}{Tabs.}

\def\cvprPaperID{8701} 
\def\confName{CVPR}
\def\confYear{2023}

\begin{document}

\title{IncepFormer: Efficient Inception Transformer with Pyramid Pooling \\for Semantic Segmentation}

\author{Lihua Fu$^1$ \quad 
Haoyue Tian$^1$ \quad 
Xiangping Bryce Zhai$^1$ \thanks{Corresponding author.}  \quad  
Pan Gao $^{1 \ *}$ \quad 
Xiaojiang Peng$^2$ \\
$^1$College of Computer Science and Technology, Nanjing University of Aeronautics and Astronautics\\
$^2$College of Big Data and Internet, Shenzhen Technology University
}

\maketitle

\begin{abstract}
Semantic segmentation usually benefits from global contexts, fine localisation information, multi-scale features, etc. To advance Transformer-based segmenters with these aspects, we present a simple yet powerful semantic segmentation architecture, termed as IncepFormer. IncepFormer has two critical contributions as following. First, it introduces a novel pyramid structured Transformer encoder which harvests global context and fine localisation features simultaneously. These features are concatenated and fed into a convolution layer for final per-pixel prediction. Second, IncepFormer integrates an Inception-like architecture with depth-wise convolutions, and a light-weight feed-forward module in each self-attention layer, efficiently obtaining rich local multi-scale object features. Extensive experiments on five benchmarks show that our IncepFormer is superior to state-of-the-art methods in both accuracy and speed, \textit{e.g.}, 1) our IncepFormer-S achieves 47.7\% mIoU on ADE20K which outperforms the existing best method by 1\% while only costs half parameters and less FLOPs. 2) Our IncepFormer-B finally achieves 82.0\% mIoU on Cityscapes dataset with 39.6M parameters. Code is available:\href{https://github.com/shendu0321/IncepFormer}{github.com/shendu0321/IncepFormer}

\end{abstract}

\section{Introduction}
\label{sec:intro}
Semantic segmentation, as one of the most fundamental and challenging research topics in computer vision, with a wide range of applications, including autonomous 
driving, robotics and medical imaging, has attracted substantial attention over the past decades. It aims at assigning each pixel a semantic category, 
thus different from image classification, image-level prediction. 

Recent methods in semantic segmentation are usually  based on an encoder-decoder architecture where the encoder generates feature maps by downsampling, and 
the decoder upsamples feature maps to high-resolution segmentation mask with per-piexl category scores. The early CNN-based models, including the representative
FCN~\cite{long2015fully} and DeepLab family \cite{chen2014semantic,chen2017deeplab,chen2017rethinking,chen2018encoder}, capture the rich semantic information through 
convolution and the variants of convolution, such as dilated convolution, 
achieving state-of-the-art results on semantic segmentation task. Gradually, with the great progress in natural language proceeding (NLP), there has been increasing interest 
to employ Transformers to vision tasks. Dosovitskiy \etal~\cite{dosovitskiy2020image} proposes vision Transformer (ViT) for image classification, which is considered as
the first work of the application of Transformer to the vision domain. In ViT, they
split an image into a sequence of embedding patches (tokens) and update the patch features progressively via self-attention, 
leading to a classification performance boost on ImageNet. Inspired by this work, Zheng \etal~\cite{zheng2021rethinking} developes the SETR to show the superiority of using ViT as the backbone 
in semantic segmentation task.

\begin{figure}[t]
  \centering
   \includegraphics[width=1.0\linewidth]{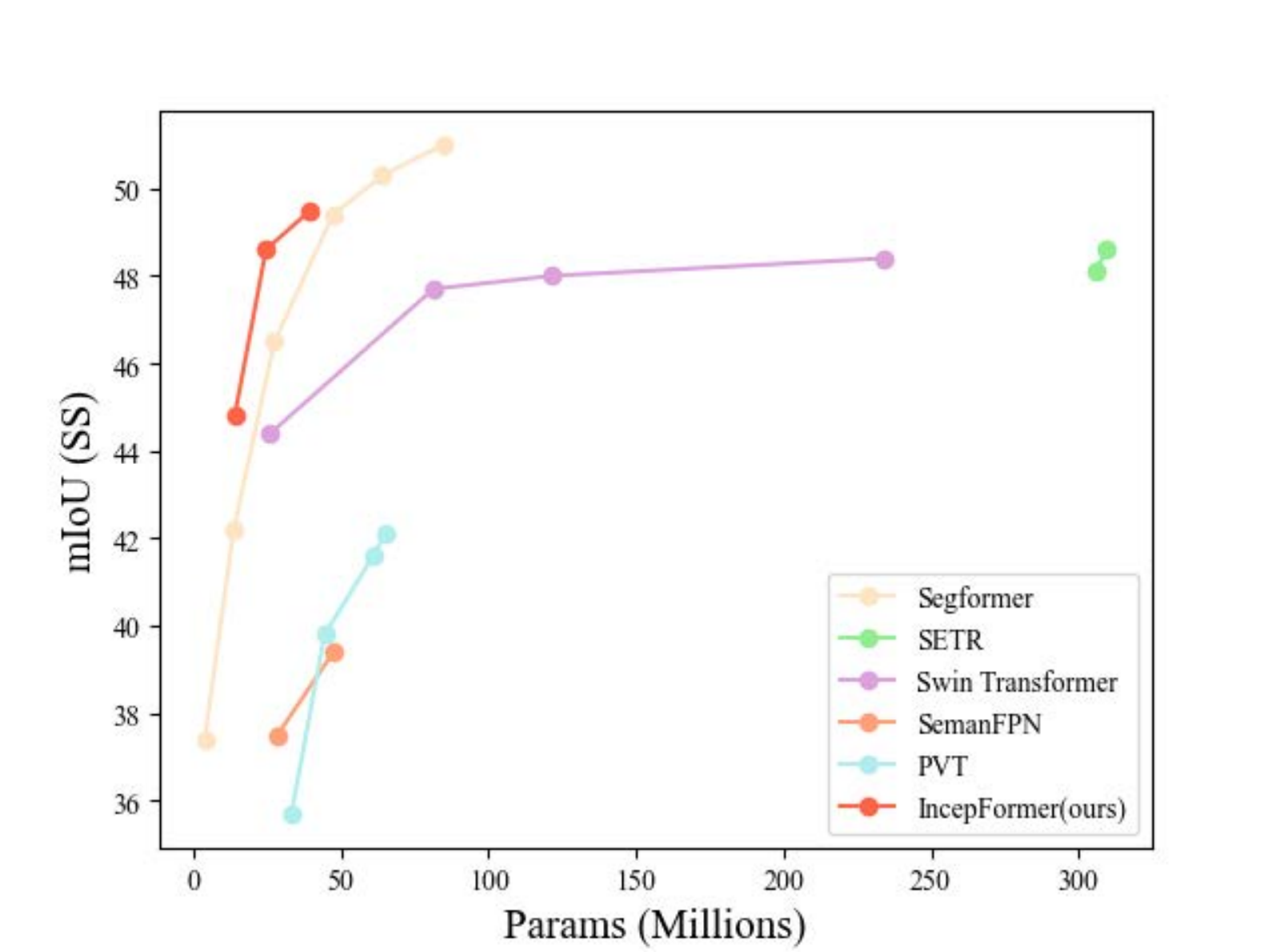}
   \caption{Performance-Parameters curves on the ADE20K validation sets. All results are reported with a single model and tested by single-scale data.
   We can see that our IncepFormer achieves the best balance between performance and parameters compared with previous methods.}
   \vspace{-3mm}
   \label{fig:F1}
\end{figure}

Despite the good performance achieved by SETR, ViT has some limitations: 1) The standard self-attention mechanism brings a huge amount of computational complexity that is 
the quadratic with regard to the number of input tokens. 2) The output feature maps of the ViT are single-scale, which may render the features extracted lack rich contextual information. 
To address these problems, state-of-the-art Transformer models resort to  downsampling strategies to reduce the feature size and thus design a hierarchical encoder architecture. 
Pyramid Vision Transformer (PVT)~\cite{wang2021pyramid} is the first work to employ the pyramid structure into dense prediction. After that, 
Xie \etal~\cite{xie2021segformer} proposes Mix Transformer (MiT), which is also a pyramid structure, showing the considerable improvement over the SETR counterpart on semantic segmentation.
Swin Transformer, another popular hierarchical vision transformer, calculates the self-attention in a local window and yields linear complexity with image size.
However, these methods only consider the multi-scale nature across stages/layers, and overlooked the multi-scale nature of objects within one attention layer, i.e., \emph{in-self-attention multi-scale},
resulting in the incapacity to capture rich features  in different sized objects.

To address the above limitations, we introduce Efficient Inception Transformers with Pyramid Pooling for Semantic Segmentation (IncepFormer), a novel and universal Transformer framework.
First, at the encoder, 
we propose a self-attention mechanism, termed Inception Multi-Head Self-Attention (Incep-MHSA). Unlike traditional self-attention that merge tokens 
purely relying on dot-product, we integrate the convolution and pooling into self-attention, which effectively induces inductive bias into feature learning. We apply an eclectic convolution
kernel size to adjust the receptive field flexibly according to the shape of the objects in segmentation scenes. In addition, 
we change the norm layer to BatchNorm, as the semantic segmentation  orients the network towards 2D image rather than 1D sequence. Further, we propose an efficient fead-forward 
network (E-FFN). As convolution has natural location information, we discard the position embedding.
Secondly, for the decoder, 
we design a lightweight Upsample-Concat decoder where the core idea is to make full use of the encoder-generated features. That is, the features of lower stages 
tend to local information, while the ones of higher stages tend to global information. By integrating the features from different stages, the Upsample-Concat 
decoder merges the local and global attention. As a result, we obtain a simple but powerful encoder-decoder architecture that can adapt to random test resolution 
without impairing the performance, demonstrating the great potentials.  

\indent{Our contributions can be summarised as follows:}  
\begin{itemize}
  \item A pyramid transformer encoder, which not only considers the multi-scale in feature maps across stages, but also incorporates the multi-scale nature inside self-attention mechanism via an inception-like architecture. 
  \vspace{-2mm}
  \item A simple but powerful Upsample-Concat decoder that merges fine localization and global context information with extremely low computational cost.
  \vspace{-2mm}
  \item We design three different sized versions for our proposed IncepFormer. As shown in \cref{fig:F1}, our IncepFormer outperforms recent famous and classical methods and achieves the best trade-off between performance and computational cost on ADE20K dataset.
\end{itemize}

\section{Related Work}
\noindent\textbf{Semantic Segmentation.} Semantic segmentation is a challenging and fundamental computer vision task. The fully convolution network (FCN), 
the most representative CNN-based model, is applied to pixel-wise predictions by removing the fully connected layers. After that, the researchers make 
many efforts from various perspectives to address the limited receptive field/context modeling problems in FCN. PSPNet~\cite{zhao2017pyramid} introduces the PPM module 
to get contextual information of different regions 
while DeepLabV2~\cite{chen2017deeplab} develops ASPP module which employs pyramid dilated convolutions with different dilated rates. Besides, attention-based models 
are popular for capturing long-range contextual information. DANet~\cite{fu2019dual} adopts both channel attention and spatial attention. CCNet~\cite{huang2019ccnet} 
alternatively pays attention to saving the massive computational budget introduced by full-spatial attention. These methods significantly improve semantic 
segmentation performance based on classical CNN-based networks like VGG~\cite{simonyan2014very} and ResNet~\cite{he2016deep}. Recently, 
some methods \cite{zheng2021rethinking, xie2021segformer} have shown the great potential of the Transformer-based architectures in this area.

\noindent\textbf{Encoder-Decoder Architectures.} In the field of deep learning, the framework of semantic segmentation usually consists of two parts: encoder and decoder. 
For the encoder, researches used to adopt the popular classification networks, transfering them to the downstream tasks. However, they may not perform well, since semantic 
segmentation is a kind of dense prediction compared to image level prediction. Hence, the tailored encoders 
are designed, represented by SETR~\cite{zheng2021rethinking} and SegFormer~\cite{xie2021segformer}. SETR~\cite{zheng2021rethinking} adopts ViT~\cite{dosovitskiy2020image} 
as the backbone to extract features, achieving excellent performance, while SegFormer~\cite{xie2021segformer} proposed a hierarchical transformer encoder 
which outputs multi-scale features. For the decoder, it is often used to aggregate features from different encoder layers or stages. 
Decoders are often designed to achieve the goals, including combing both local context and global context, and unifying multi-scale 
semantics.

\begin{figure*}[t]
  \centering
    \includegraphics[width=1.0\linewidth]{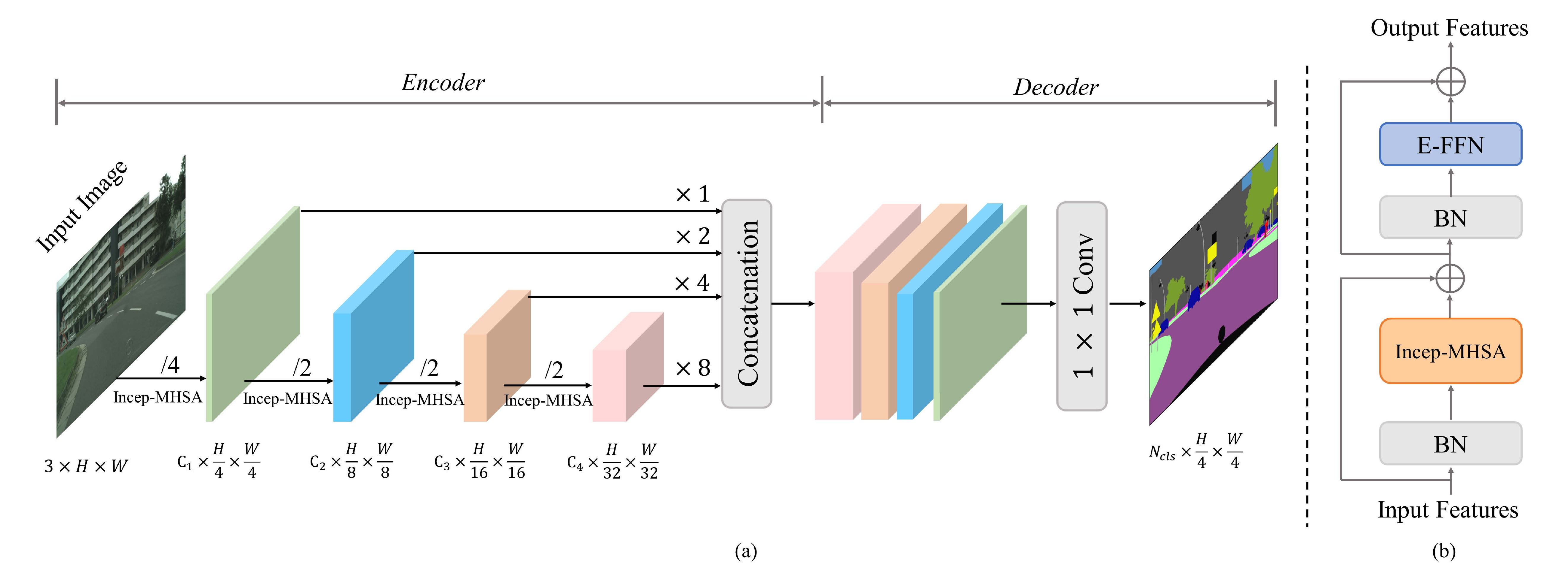}

    \caption{\textbf{(a): The proposed IncepFormer architecture}, which consists of two main parts: 1) A pyramid transformer encoder to capture the coarse and fine-grained features; 
    and 2) a lightweight Upsample-Concat decoder to diametrically merge the multi-scale features and do the peixl-level prediction. ``$/n$'' means that the H and W of the 
    feature map is reduced by a factor $n$ when downsampling. ``$\times n$'' indicates $n$ times upsampling. \textbf{(b):} Details of our Inception Transformer Block (IPTB)}
    \label{fig:incepformer}
\end{figure*}

\noindent\textbf{Efficient Self-Attention Mechanism.} Starting from ViT, more and more works introduce
the self-attention in the tasks of computer vision. Since the standard self-attention layer costs a huge computation resource for high resolution
image, two popular strategies appear: 1) divide the features into regions, and perform the local self-attention within the regions or 2) merge the tokens
to decrease the number of the tokens. A typical work of the local self-attention is Swin Transformer~\cite{liu2021swin} that divides the features into 
several regions via shifted window and does the self-attention in each region separately. Regarding to the token merging, 
the relative works are PVT series \cite{wang2021pyramid,wang2022pvt}, which design a spatial-reduced attention to merge tokens of key and value. The PVT and other analogous
methods not only reduce the computational complexity, but also can remove fixed size position embedding due to the use of convolution, demonstrating the potential of Transformer backbone 
compared to the CNN-based counterparts in dense prediction tasks.

However, either PVT or SegFormer reduces the spatial scale of $\textbf{K}$ and $\textbf{V}$ before the attention operation only through a single branch
such as convolution, resulting in the simplification of the features collectd and the loss of fine-grained information of small objects. Therefore, we put forward the Incep-MHSA
mechanism that can preserve overall and detailed visual knowledge.

\section{Method}
In this section, we first provide an overview of our IncepFormer networks in \cref{method-overview}. Then, we present the architecture of IncepFormer based on multi-scale 
convolution-and-pooling MHSA in \cref{method-encoder} and the proposed lightweight Upsample-Concat decoder in \cref{method-decoder} respectively.

\subsection{Overview} \label{method-overview}
As shown in \cref{fig:incepformer}(a), IncepFormer consists of two main parts: 1) A pyramid Inception Transformer encoder to capture the coarse and fine-grained features; 
and 2) a lightweight Upsample-Concat decoder to diametrically merge the multi-scale features and do the piexl-level prediction.

Given an image of size $H \times W \times 3$, firstly we divide it into patches of size $4 \times 4$. Unlike ViT using patches of size $16 \times 16$, smaller patches
is conducive to intense prediction tasks. Then we use these patches as the input of the hierarchical Transformer encoder, obtaining the multi-level features
at $\left\{ \frac{1}{4}, \frac{1}{8}, \frac{1}{16}, \frac{1}{32} \right\}$ of the original image resolution. 
The pyramid pooling across stages is done using the same patching merging as in the previous works PVT and SegFormer. 
Next, we deliver these multi-scale features to the Upsample-Concat
decoder to generate the segmentation mask at the size $N_{cls} \times \frac{H}{4} \times \frac{W}{4}$, where $N_{cls}$ is the number of categories. In the rest of the 
section, we will describe the details of encoder and decoder.

\subsection{Inception Transformer Encoder} \label{method-encoder}
Convolution or pooling is widely used in CNN network design in various computer vision tasks. Recently, the authors of \cite{xie2021segformer, wu2022p2t} applied 
them in self-attention layer to reduce the 
computational complexity. However, existing literature usually applies one of them, which lack the features diversity. 
To this end, we explore the combination of convolution and pooling in transformers backbone networks, targeting at improving the semantic segmentation 
tasks generally. Further, we adapt the idea of multi-scale convolution in InceptionNet~\cite{szegedy2016rethinking} to the transformer. 
With the above consideration, the proposed Inception transformer can capture richer contextual information while reducing the computational complexity significantly. 

Let us introduce the Inception Transformer Block (IPTB), the structure of which is illustrated in \cref{fig:incepformer}(b). The input is first inputted into a normalization layer, and then
passed into Incep-MHSA, whose output is residual-connected
with the original input. Here, we use BatchNorm instead of LayerNorm used in vanilla transformer for better adapting to 2D image structure.
In the second sub-block, we make fine adjustments to the feed-forward network, named E-FFN, according
to the characteristics of the 2D-image, for feature projection. Similarly, BatchNorm is applied before the sub-block, and residual connection is applied after the sub-block. In summary, the above process can be formulated as:
\begin{equation} \label{eqn1}
  \begin{split}
  & \textbf{X}_{att} = {\rm \textbf{X} + Incep\mbox{-}MHSA \left( BatchNorm \left(\textbf{X}\right) \right) },\\
  & \textbf{X}_{out} = \textbf{X}_{att} + {\rm E\mbox{-}FFN } \left( {\rm BatchNorm } \left( \textbf{X}_{att} \right) \right),
  \end{split}
\end{equation}
where $ \textbf{X}, \textbf{X}_{att},$ and $ \textbf{X}_{out}$ are the input, the output of Incep-MHSA, and the output of the inception transformer block, respectively.

\begin{figure}[t]
  \centering
    \includegraphics[width=1.0\linewidth]{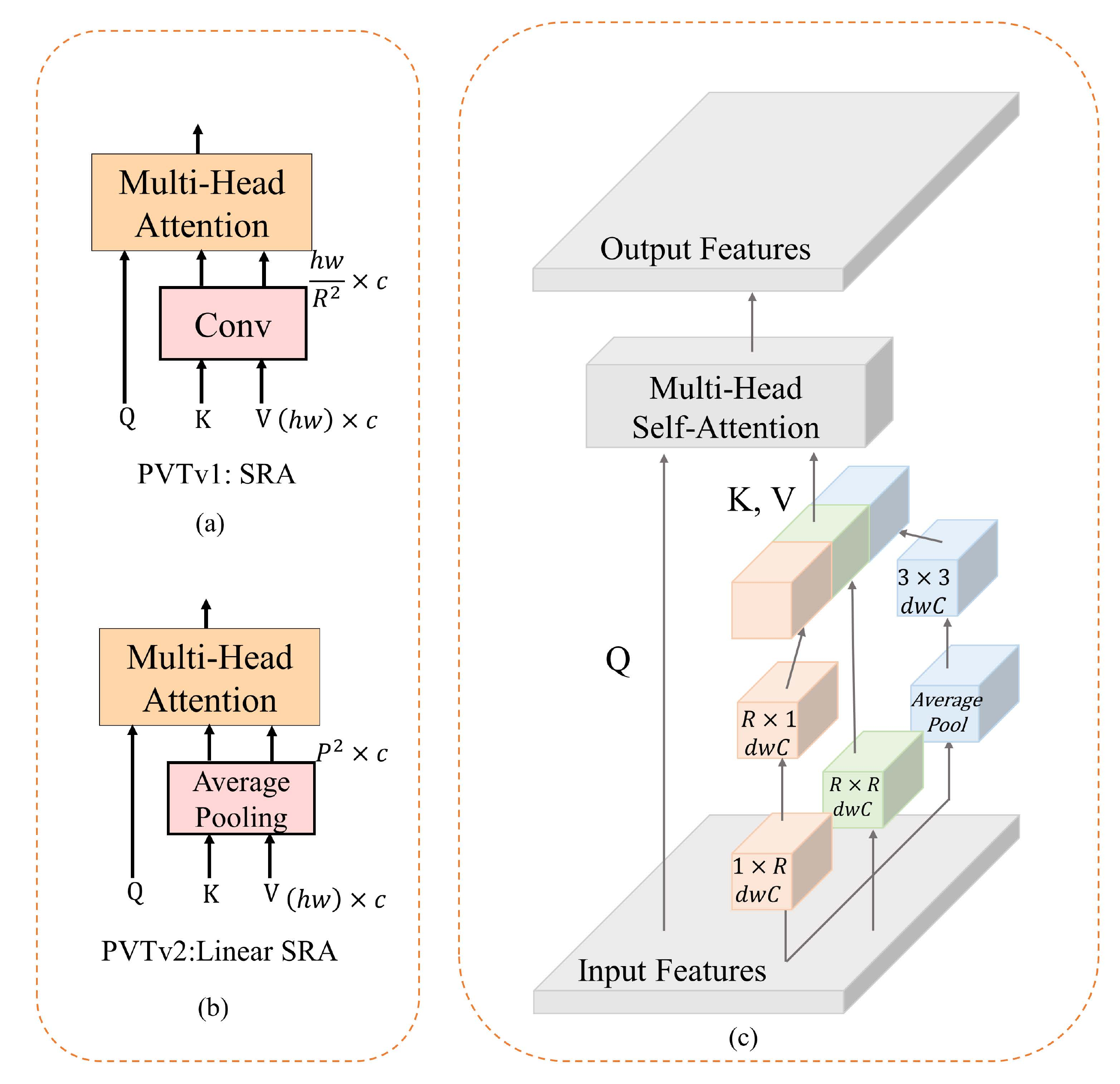}

    \caption{Comparison of SRA in PVT v1, linear SRA in PVT v2 and our proposed Incep-MHSA. In (c), ``$k_1 \times k_2\, dwC$'' means a depth-wise convolution (${\rm dwC}$) using the kernel size of $k_1 \times k_2$.
    $R$ is the ratio to reduce the resolution of the features.}
    \label{fig:Incep-MHSA}
\end{figure}

\begin{figure}[t]
  \centering
    \includegraphics[width=0.8\linewidth]{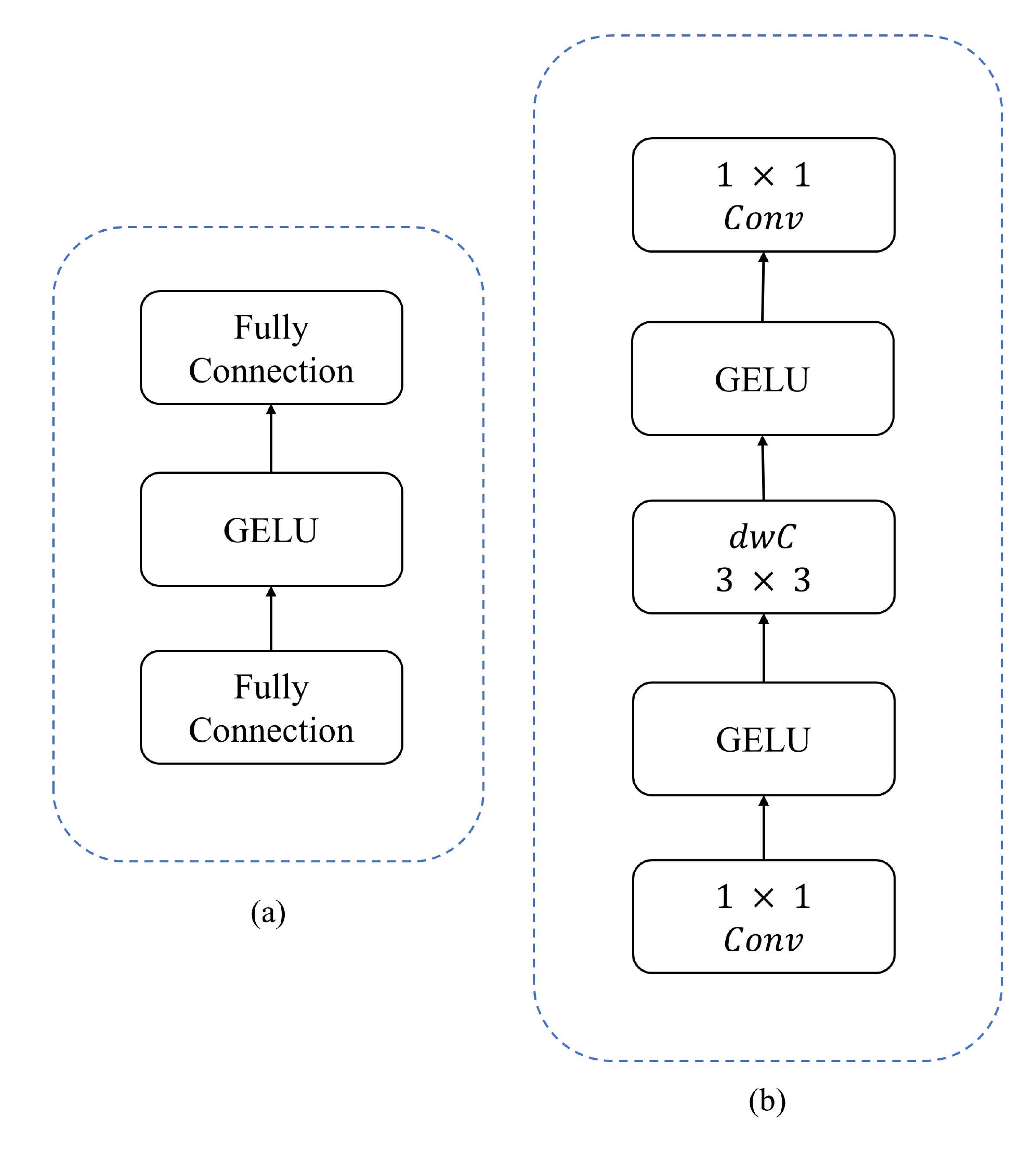}
    \caption{Comparison of the original feed-forward network and our efficient feed-forward network (E-FFN)}
    \label{fig:E-FFN}
\end{figure}

\subsubsection{Incep-MHSA}
Here, we explain the design of our Incep-MHSA. As shown in \cref{fig:Incep-MHSA}, different from SRA and  linear SRA which use convolutions and average pooling separately for spatial reduction
before attention operation, we apply three different branches on $\textbf{X}$ to generate feature maps inspired by Inception~\cite{Szegedy_2016_CVPR}. In \cref{fig:Incep-MHSA}(c), at the first branch,
$\textbf{X}$ goes through a depth-wise convolution with the kernel size of $1 \times R$, then again depth-wise convolution with kernel size of $R \times 1$. The second branch is applied with a 
depth-wise convolution, whose kernel size is $3 \times 3$, on the $\textbf{X}$. The third branch firstly uses an average pooling with reduction ratio $R$, and then use $3\times3$ depth-wise convolution. 
Note that, for the former two branches, we set the stride for the convolution as $R$, and thus their outputs are also down-sampled by $R\times R$ spatially.
In summary, the above process is formulated as 
\begin{equation} \label{eqn2}
  \begin{split}
  {\rm \textbf{C}_{1}} &= {\rm dwC}_{R \times 1} \left({\rm dwC}_{1 \times R} \left( \textbf{X} \right)\right), \\
  {\rm \textbf{C}_{2}} &= {\rm dwC}^3 \left( \textbf{X} \right), \\
  {\rm \textbf{P}_{3}} &= {\rm dwC}^3 \left({\rm AvgPool } \left( \textbf{X} \right)\right), \\
  \end{split}
\end{equation}
where $\left\{\rm \textbf{C}_{1}, \textbf{C}_{2}, \textbf{P}_{3}\right\}$ denotes the outputs of three branches respectively, and ${\rm dwC}_{k_1 \times k_2} \left( \cdot \right)$ or $ {\rm dwC}^n$ indicates the depth-wise 
convolution with the kernel size of $k_1 \times k_2$ or $n \times n$ respectively. After that, we flatten and concatenate these features maps:
\begin{equation} \label{eqn3}
  {\rm\textbf{O}} = {\rm LayerNorm(Concat(\textbf{C}_{1}, \textbf{C}_{2}, \textbf{P}_{3}))},
\end{equation}
where, for simplicity, the flattening operation is omitted. In this way, the obtained token sequence ${\rm\textbf{O}}$ is shorter than the input $\textbf{X}$ flattened. Moreover, $\rm\textbf{O}$ contains rich contextual
abstraction of the input $\textbf{X}$ and can thus act as the alternative of the input $\textbf{X}$ when computing MHSA.

\begin{figure*}[t]
  \centering
    \includegraphics[width=1.0\linewidth]{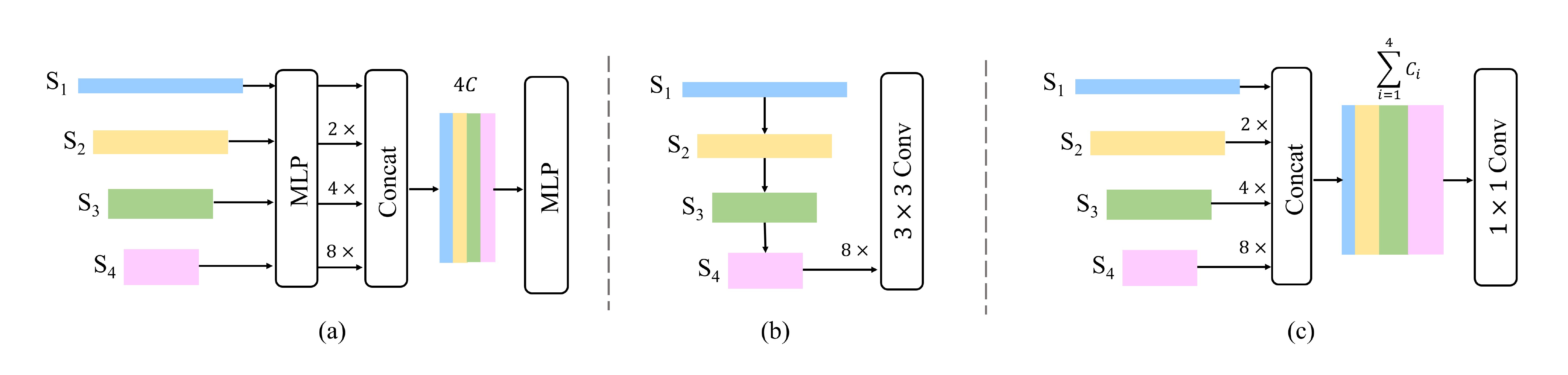}
    \caption{Three different encoder architectures. $S_i$ means the stage $i$.}
    \label{fig:decoder designs}
\end{figure*}

Denote the query, key and value tensors in MHSA by $\textbf{Q}, \textbf{K}$ and $\textbf{V}$, respectively. We replace the traditional
\begin{equation} \label{eqn4}
  \left( \textbf{Q}, \textbf{K}, \textbf{V} \right) = \left( \textbf{XW}^q, \textbf{XW}^k, \textbf{XW}^v \right), 
\end{equation}
with 
\begin{equation} \label{eqn5}
  \left( \textbf{Q}, \textbf{K}, \textbf{V} \right) = {\textbf{XW}^q, \textbf{OW}^k, \textbf{OW}^v},
\end{equation}
in which $\textbf{W}^q, \textbf{W}^k$ and $\textbf{W}^v$ represent the weight matrix of linear transformation used to generate query, key and value tensors respectively. Thus, $\textbf{Q}, \textbf{K}, \textbf{V}$
are fed into the self-attention module to compute the attention, which is formulated as follows:
\begin{equation} \label{eqn6}
  {\rm Attention} = \rm{Softmax \left(\frac{\textbf{QK}^\textbf{T}}{\sqrt{d_k}} \right)}\textbf{V},
\end{equation}
where $d_k$ is the channel dimension of $\textbf{K}$, and division by $\sqrt{d_k}$ can be considered as an approximate normalization. The Softmax function is applied along each row in the matrix.
\cref{eqn6} can be easily generalized to multi-head self-attention case. It is worth pointing out that, using depth-wise convolution for Inception transformer not only can reduce the computation complexity,
but also can increase its channel attention modeling capability for the key and value features. This supplements the self-attention calculation in \cref{eqn6}, which computes the attention weight 
between two feature vectors using dot-product attention and ignores the channel attention among individual channels \cite{zhao2021point,chen2022mixformer}.

\subsubsection{E-FFN}
There is no doubt that the feed-forward network (FFN) is a significant component in the transformer block for feature enhancement. As described in \cref{fig:E-FFN}(a), most of the previous methods follow 
the original FFN, in vanilla transformer applied in NLP area. Though effective, this design is not capable at learning 2D locality, which is, however, indispensable for semantic segmentation task. 
Then the SegFormer~\cite{xie2021segformer} proposed Mix-FFN, where a $3 \times 3$ depth-wise convolution is inserted between the first fully-connected (FC) layer and GELU. However, this needs to frequently 
switch between 1D sequence and 2D image when implementing, incurring a large amount of computational cost. To this end, we replace the FC layer with $1 \times 1$ convolution, so that the generated FFN network can 
inherit the advantages of the CNN (\ie 2D locality).

First, we transform the input sequence ${\textbf{X}_{att}}$ to a 2D feature map ${\textbf{X}_{in}}$:
\begin{equation} \label{eqn7}
  \begin{aligned}
    \textbf{X}_{in} &=  {\rm Seq2Img} \left( \textbf{X}_{att}\right), \\
    \textbf{X}_{out} &= {\rm Conv}^1 \left({\rm dwC}^3 \left ({\rm Conv}^1 \left ({\rm BN}( \textbf{X}_{in}) \right) \right) \right) + \textbf{X}_{in}, \\
  \end{aligned}
\end{equation}
where ${\rm Seq2Img} \left( \cdot \right)$ means reshaping the 1D sequence to 2D feature map. $ {\rm Conv}^n$ denotes the common convolution with the kernel size of $n \times n$.
${\rm dwC}^n$ means the depth-wise convolution with $n \times n $ kernel size similarly.

With the above defined Icep-MHSA and E-FFN, we design three encoder models with different size, termed as tiny version, IPT-T, small version, IPT-S, and base version, IPT-B, respectively. 
The corresponding segmentation framework are named as IncepFormer-T,
IncepFormer-S, IncepFormer-B, respectively. Detailed network settings are presented in supplementary materials.

\subsection{A Simple Upsample-Concat Decoder} \label{method-decoder}
The proposed Upsample-Concat decoder contains three steps.
First, feature maps at each stage are upsampled to $\frac{1}{4}$th and concatenated together. Then, a $1 \times 1$ convolution
is adopted to linearly transform the concatenated feature map. Finally, the concatenated feature map is fed into another $1 \times 1$ convolution to predict the segmentation mask M, whose
resolution is $\frac{H}{4} \times \frac{W}{4} \times N_{cls}$, in which $N_{cls}$ is the number of categories.
The process can be formulated as:
\begin{equation} \label{eqn8}
  \begin{aligned}
    &\hat{F}_i = {\rm Upsample}\left(\frac{H}{4} \times \frac{W}{4} \right)\left( F_i \right), \\
    &F = {\rm Conv^1}\left( \sum_{i=0}^{4}C_{\hat{F}_i},C\right) \left({\rm Concat} \left(\hat{F}_i \right) \right), \\
    &M = {\rm Conv^1} \left(C, N_{cls} \right) \left( F \right), \quad \forall i \in \ \left\{ 1,2,3,4 \right\} \\
  \end{aligned}
\end{equation}
where $M$ represents the predicted segmentation mask, and ${\rm Conv^1}(C_{in}, C_{out})(\cdot)$ refers to a $1 \times 1$ convention with $C_{in}$ and $C_{out}$ as input and output channels,
respectively.

Obviously, our decoder is simple but powerful that drops the component with high computing requirements compared with other typical methods. Of course, the key to drive 
this decoder is that our pyramid Transformer encoder, which not only has  rich contextual information and effective receptive field (ERF), but also considers the characteristics of small objects
by  using multiple strip-shaped convolutional kernels.

\begin{table}[t]
  \centering
  \begin{tabular}{c|c|c|c}
  \toprule[1pt]
    Method & Params(M) & GFLOPs & Top-1(\%)  \\
    \midrule
    PVT-T~\cite{wang2021pyramid} & 13.2 & 1.9  & 75.1   \\
    MiT-B1~\cite{xie2021segformer} & 13.4 & 2.1 & 78.7  \\ 
    IPT-T & 14.0 & 2.3 & \textbf{80.5}  \\ 
    \midrule
    PVT-S~\cite{wang2021pyramid}  & 24.5 & 3.8 & 79.8  \\
    MiT-B2~\cite{xie2021segformer} & 25.4 & 4.0 & 82.0  \\
    Swin-T~\cite{liu2021swin} & 28.3 & 4.5 & 81.3  \\
    ConvNeXt-T~\cite{liu2022convnet} & 28.6 & 4.5 & 82.1  \\
    IPT-S & 24.3 & 4.7 & \textbf{82.9}  \\
    \midrule
    PVT-M~\cite{wang2021pyramid}  & 44.2 & 6.7 & 81.2  \\
    MiT-B3~\cite{xie2021segformer} & 45.2 & 6.9 & 83.2  \\
    Swin-S~\cite{liu2021swin} & 49.6 & 8.7 & 83.0  \\
    ConvNeXt-S~\cite{liu2022convnet} & 50.1 & 8.7  & 83.1  \\
    IPT-B & 39.3 & 7.8 & \textbf{83.6}  \\
  \bottomrule[1pt]
  \end{tabular}
  \caption{Comparison with state-of-the-art methods on ImageNet validation set. ``Top-1(\%)'' denotes Top-1 accuracy.}
  \label{tabel:imageNet}
  \vspace{-2mm}
 \end{table}

\section{Experiments}
\noindent\textbf{Datasets.} We evaluate our methods on six widely-used datasets, including one image classification dataset and five semantic segmentation benchmark datasets.
ImageNet-1K~\cite{deng2009imagenet} is a well-known dataset for image classification containing 1,000 categories. We pretrain IPT encoder on this dataset as done in most segmentation methods.
ADE20K~\cite{zhou2017scene} and COCO-Stuff~\cite{caesar2018coco} are both challenging datasets. The former, with 150 semantic classes, consists of 20,210/2,000/3,352 images in the training, validation 
and test sets, respectively. The latter covers 172 semantic concepts, made up of 164k images. Cityscapes~\cite{cordts2016cityscapes} is an urban scenes dataset and contains 5,000 high-resolution images 
with 19 classes. There are 2,975/500/1,525 images for training, validation and testing, separately. Pascal VOC~\cite{everingham2010pascal} has 20 foreground classes 
and a background class, where it has 10, 582/1, 449/1, 456 images for the three set split in this dataset. Pascal Context~\cite{mottaghi2014role} contains 59 foreground classes and a background 
class, whose training set and validation set contain 4,996 and 5,104 images, respectively.

\noindent\textbf{Metrics.} As usual, we adopt Top-1 accuracy and mean Intersection over Union (mIoU) as the evaluation metrics for classiﬁcation and segmentation, respectively.

\noindent\textbf{Implementation details.} We use timm~\cite{rw2019timm} and mmsegmentation~\cite{mmseg2020} libraries to implement our methods for classification and segmentation tasks, respectively. The encoder 
module of our IncepFormer architecture is pretrained on ImageNet-1K dataset. All models are trained using 8 RTX 3090 GPUs.

For ImageNet-1K pretraining, the implementation setting is the same as DeiT~\cite{touvron2021training}. For segmentation experiments, we apply data augmentation, including random scaling with ratio
0.5-2.0, random horizontal flipping and random cropping. We trained the models using AdamW optimizer and the initial learning rate is 0.6 with the poly-learning rate decay.
We train our model 160K iterations for ADE20K, COCO-Stuff dataset and 80K iterations for Cityscapes, Pascal Context and Pascal VOC dataset, while for the second point of ablation studies, influence of output channels $C$ on the decoder, we train the model for 40K iterations.
We set a batch size of 8 for Cityscapes dataset and 16 for the other datasets. 

\subsection{Image Classfication Results on ImageNet}
Pretraining encoder is a common strategy for training segmentation models. Here, we compare our IPT with recently developed backbones. As shown in \cref{tabel:imageNet},
our IPT achieves superior results compared with the popular transformer-based models, like PVT, Swin-Transformer and MiT, the encoder of SegFormer, and also outperforms
the recent start-of-the-art CNN-based backbone, ConvNeXt~\cite{liu2022convnet}.

\subsection{Ablation study}
\noindent\textbf{Efficiency of the proposed model.} \Cref{fig:F1} shows the relationship of performance, versus encoder size on ADE20K dataset and the results of three versions of our proposed model with three sizes on four datasets can be found in supplementary materials. The first phenomenon can be observed is that our decoder is sufficiently lightweight that the parameters of
all decoders are less than 1M. Besides, we can conclude that, with the increasing of the size of the encoders, consistent improvements are yielded on all datasets.

\noindent\textbf{Influence of output channels $C$ on the decoder.} This part explores the influence of the output channels of $1 \times 1$ Conv on the decoder module, see
\cref{method-decoder}. In \cref{tabel:ablation C}, we show the performance, flops and parameters in different output channels of $1 \times 1$ Conv. As $C$ increases, the performance 
improves. However, it also leads to the larger parameters and slower efficiency. The performance seems to plateaus for the output channels larger than 768. So we choose $C$ = 512 for IncepFormer-T
and $C$ = 768 for the rest.

\begin{figure*}[t]
  \centering
    \includegraphics[width=1.0\linewidth]{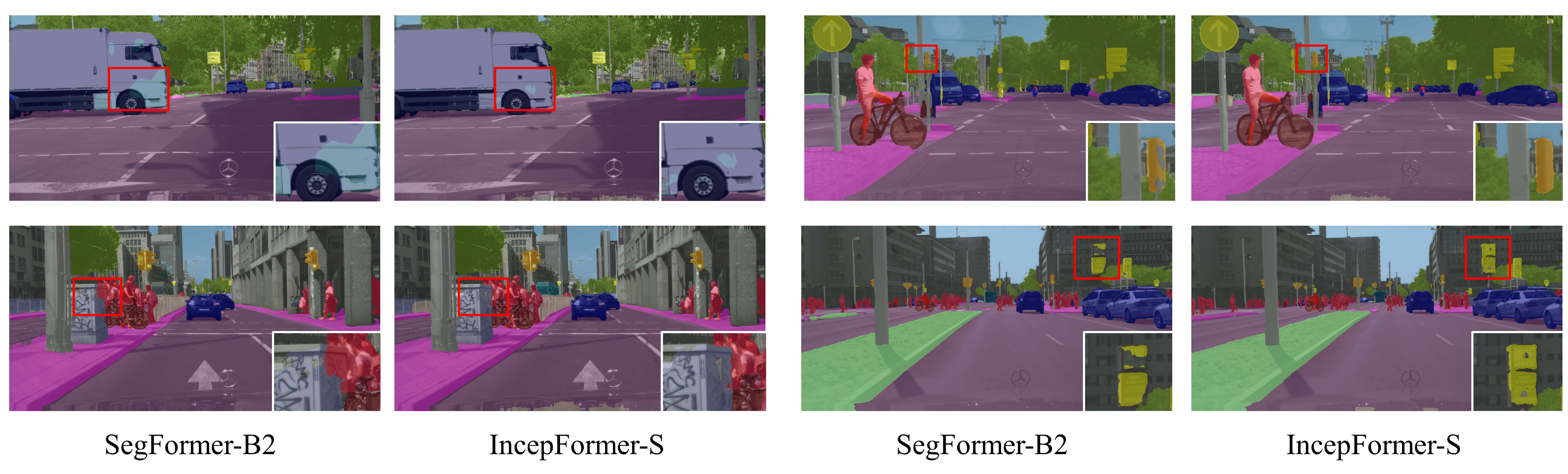}
    \caption{Qualitative comparison of   SegFormer-B2 and IncepFormer-S on the Cityscapes dataset. More visual results can be found in supplementary material.}
    \label{figure:compare figure}
    \vspace{-2mm}
\end{figure*}

\begin{table}[t]
  \centering
  \renewcommand{\arraystretch}{1.4}
  \begin{tabular}{c|c|c|c|c}
    \toprule[1pt]
      $C$ & Params(M) & GFLOPs & \multicolumn{2}{c}{mIoU(SS/MS)} \\
      \cline{1-5}
      256	& 24.1 & 29.3 & 42.1 & 43.3 \\
      \cline{1-5}
      512	& 24.4 & 33.9 &	42.3 & 43.8 \\
      \cline{1-5}
      768	& 24.6 & 38.5 &	42.5 & 43.8 \\
      \cline{1-5}
      1024 & 25.0 &	43.1 & 42.6 &	44.0 \\
      \cline{1-5} 
      2048 & 26.1 &	61.4 & 42.7 &	44.1 \\
    \bottomrule[1pt]
    \end{tabular}
    \caption{The influence of output channels $C$ on the decoder on ADE20K.}
    \label{tabel:ablation C}
    \vspace{-2mm}
\end{table}

\begin{table}[t]
  \centering
  \renewcommand{\arraystretch}{1.2}
  \scalebox{0.8}{
  \begin{tabular}{c|c|c|c|c}
   \toprule[1pt]
    Structure & Params(M) & GFLOPs & \multicolumn{2}{c}{mIoU(SS/MS)}  \\
    \midrule
    IncepFormer-T(a) & 15.8 & 32.4 & 44.2 & 45.7 \\
    \midrule
    IncepFormer-T(b) & 16.6 & 16.9 & 41.7 & 44.3 \\
    \midrule
    IncepFormer-T(c) & 14.0 & 21.2 & 44.8 & 46.3 \\
    \midrule
    IncepFormer-T(c) w/o S$_1$ & 13.9 & 20.7 & 44.6 & 46.2  \\
   \bottomrule[1pt]
  \end{tabular}}
  \caption{Results on different encoder designs. IncepFormer-T(a) means \cref{fig:decoder designs} is used as decoder. IncepFormer-T(c) w/o S$_1$ denotes without the output
  of stage 1 in the decoder. The number of parameters and flops are calculated in $512 \times 512$.}
  \label{table:encoder designs}
  \vspace{-3.5mm}
 \end{table}

 \begin{table*}[t]
  \centering
  \resizebox{\textwidth}{!}{
  \begin{tabular}{c|c|c|c|c|c|c|c|c|c|c|c}
    \toprule[1pt]
    \multirow{2}*{Model} & \multirow{2}*{Backbone} & Params & \multicolumn{3}{c|}{ADE20K} & \multicolumn{3}{c|}{Cityscapes} & \multicolumn{3}{c}{COCO-Stuff}\\
    \cline{4-12}
    & & (M)& GFLOPs & \multicolumn{2}{c|}{mIoU(SS/MS)} & GFLOPs & \multicolumn{2}{c|}{mIoU(SS/MS)} & GFLOPs & \multicolumn{2}{c}{mIoU(SS/MS)} \\
    \midrule
    Segformer-B1~\cite{xie2021segformer} & MiT-B1         & 13.7 & 15.9 & 42.2 & 43.1 & 243.7 & 78.5 & 80.0 & 15.9 & 40.2 & -\\
    HRFormer-S~\cite{yuan2110hrformer}   & HR Transformer & 13.5 & 109.5 & 44.0 & 45.1 & 835.7 & 80.0 & 81.0 & 109.5 & 37.9 & 38.9 \\
    IncepFormer-T  & IPT-T         & 14.0 & 21.2 & \textbf{44.8} & \textbf{46.3} & 45.4 & \textbf{80.5} & \textbf{81.8} & 21.4 & \textbf{43.8} & \textbf{44.0} \\
    \midrule
    Segformer-B2~\cite{xie2021segformer} & MiT-B2 & 27.5 & 62.4 & 46.5 & 47.5 & 717.1 & 81.0 & 82.2 & 62.4 & 44.6 & - \\
    Mask2Former~\cite{cheng2022masked} & Swin-T & 42.0 & 55.0 & 46.7 & 48.8 & - & - & - & - & - & - \\
    IncepFormer-S & IPT-S & 24.6 & 38.5 & \textbf{47.7} & \textbf{49.0} & 83.2 & \textbf{81.6} & \textbf{82.6} & 38.7 & \textbf{45.4} & - \\
    \midrule
    Segformer-B3~\cite{xie2021segformer} & MiT-B3 & 47.3 & 79.0 & \textbf{49.4} & 50.0 & 962.9 & 81.7 & 83.3 & 79 & 45.5 & - \\
    HRFormer-B~\cite{yuan2110hrformer} & HR Transformer & 56.2 & 280.0 & 48.7 & 50.0 & 2223.8 & 81.9 & 82.6 & 280.0 & 42.4 & 43.3\\
    MaskFormer~\cite{cheng2021maskformer} & Swin-T & 47.0 & 74.0 & 47.7 & 49.6 & - & - & - & - & - & - \\
    SETR-MLA~\cite{zheng2021rethinking} & ViT-Large & 310.6 & 480.7 & 48.6 & 50.1 & 588.6 & 79.3 & 82.2 & - & - & - \\
    IncepFormer-B & IPT-B & 39.6 & 54.6 & \textbf{49.4} & \textbf{50.2} & 119.3 & \textbf{82.0} & \textbf{82.7} & 54.8 & \textbf{46.4} & \textbf{46.7} \\
    \bottomrule[1pt]
  \end{tabular}}
  \vspace{-2mm}
  \caption{Comparison with state-of-the-art transformer-based methods on the ADE20K, Cityscapes and COCO-Stuff benchmark. The number of GFLOPs is calculated with the input  size of $512 \times 512$ for ADE20K and COCO-Stuff, and $768 \times 768$ for Cityscapes. ``SS/MS'' means single/multi-scale inference.}
  \label{table:transformer-based comparison}
  \vspace{-1mm}
\end{table*}

\begin{table*}[t]
  \centering
  \resizebox{\textwidth}{!}{
  \begin{tabular}{c|c|c|c|c|c|c|c|c|c|c|c}
   \toprule[1pt]
   
    \multirow{2}*{Model} & \multirow{2}*{Backbone} & Params & \multicolumn{3}{c|}{ADE20K} & \multicolumn{3}{c|}{Cityscapes} & \multicolumn{3}{c}{Pascal VOC} \\
    \cline{4-12}
    & & (M) & GFLOPs & \multicolumn{2}{c|}{mIoU(SS/MS)} & GFLOPs & \multicolumn{2}{c|}{mIoU(SS/MS)}  & GFLOPs & \multicolumn{2}{c}{mIoU(SS/MS)} \\
    \midrule
    FCN~\cite{long2015fully} & ResNet101 & 68.5 & 275.7 & 39.9 & 41.4 & 632.5 & 78.1 & 79.6 & 275.4 & 71.2 & 73.6 \\
    EncNet~\cite{zhang2018context} & ResNet101 & 54.9 & 218.8 & 42.6 & 44.0 & 501.8 & 76.1 & 77.0 & - & - & \\
    PSPNet~\cite{zhao2017pyramid} & ResNet101 & 68.0 & 256.4 & 44.4 & 45.4 & 588.2 & 79.8 & 81.8 & 256.2 & 79.8 & 81.1  \\
    CCNet~\cite{huang2019ccnet} & ResNet101 & 68.8 & 278.4 & 44.0 & 45.1 & 638.7 & 79.5 & 80.7 & 278.1 & 78.9 & 78.9\\
    DANet~\cite{fu2019dual} & ResNet101 & 68.8 & 277.7 & 44.2 & 45.0 & 635.8 & 80.5 & 82.0 & 276.9 & 76.5 & 77.3\\
    DeeplabV3~\cite{chen2017rethinking} & ResNet101 & 87.1 & 347.6 & 45.0 & 46.7 & 797.8 & 80.2 & 81.2 & 347.4 & 78.7 & 80.0\\
    DeeplabV3+~\cite{chen2018encoder} & ResNet101 & 62.6 & 255.1 & 44.1 & 45.0 & 583.2 & 80.7 & 81.5 & 254.1 & 78.6 & 79.5\\ 
    OCRNet~\cite{yuan2020object} & HRNetV2p-W48 & 70.4 & 164.8 & 43.3 & 44.9 & 324.2 & 81.4 & 82.7 &162.1 & 77.7 & 79.9 \\
    PSANet~\cite{zhao2018psanet} & ResNet101 & 73.1 & 272.5 & 43.8 & 44.8 & 637.7 & 79.7 & 80.9 & 277.5 & 77.9 & 79.3\\
    EMANet~\cite{li2019expectation} & ResNet101 & 61.1 & - & - & - & 565.2 & 79.6 & 81.0 & 246.1 & 79.6 & 81.0\\
    SemanFPN\cite{kirillov2019panoptic} & ResNet101 & 47.5 & 65.0 & 39.4 & 40.7 & 145.7 & 75.8 & 77.4 & - & - & \\
    SegNext-S~\cite{guo2022segnext} & MSCAN-T & 13.9 & 15.9 & 44.3 & 45.8 & 124.6 & 81.3 & 82.7 & 15.3 & 82.5 & -\\
    \midrule
    IncepFormer-T & IPT-T & 14.0 & 21.2 & 44.8 & 46.3 & 45.4 & 80.5 & 81.8 & 20.2 & 81.6 & 82.9\\
    IncepFormer-S & IPT-S & 24.6 & 38.5 & 48.2 & 49.2 & 83.2 & 81.6 & 82.6 & 37 & 83.1 & 83.6\\
    IncepFormer-B & IPT-B & 39.6 & 54.6 & \textbf{49.4} & \textbf{50.2} & 119.3 & \textbf{82.0} & \textbf{82.7} & 53.1 & \textbf{83.4} & \textbf{84.2} \\
   
   \bottomrule[1pt]
  \end{tabular}}
  \vspace{-2mm}
  \caption{Comparison with state-of-the-art CNN-based methods on the ADE20K, Cityscapes and Pascal VOC datasets. The input size for calculating GFLOPs is the same 
  as \cref{table:transformer-based comparison} on ADE20K and Cityscapes, and also $512 \times 512$ for Pascal VOC. }
  \label{table:cnns-based methods}
  \vspace{-2.5mm}
 \end{table*}

 \begin{table}[h]
  \centering
  \scalebox{0.85}{
  \begin{tabular}{c|c|c|c|c}
  \toprule[1pt]
  \multirow{2}*{Method} & \multirow{2}*{Backbone} & Params & \multicolumn{2}{c}{mIoU}  \\
  & & (M) & \multicolumn{2}{c}{(SS/MS)} \\
   \midrule
   FCN~\cite{long2015fully} & ResNet101 & 68.5 & 44.1 & 45.3 \\
   PSPNet~\cite{zhao2017pyramid} & ResNet101 & 68.0 & 46.0 & 47.2 \\
   DeeplabV3~\cite{chen2017rethinking} & ResNet101 & 87.1  & 46.6 & 47.8 \\
   DeeplabV3+~\cite{chen2018encoder} & ResNet101 & 62.6  & 47.2 & 48.3 \\
   EMANet~\cite{li2019expectation} & ResNet101 & 61.1  & - & 53.1 \\
   HRNet~\cite{wang2020deep} & HRNetW48 & 74.5 & - & 56.2\\
   SegNext-S~\cite{guo2022segnext} & MSCAN-T & 13.9  & 54.2 & 56.1 \\
   \midrule
   IncepFormer-T & IPT-T & 14.0 & 53.8 & 55.1 \\
   IncepFormer-S & IPT-S & 24.6  & 55.3 & 56.4 \\
   IncepFormer-B & IPT-B & 39.6 & \textbf{56.9} & \textbf{57.8} \\
  \bottomrule[1pt]
  \end{tabular}}
  \caption{Comparison on Pascal Context dataset. The number of parameters is calculated with the input size of $480 \times 480$.}
  \label{table:Pascal Context}
 \end{table}

\noindent\textbf{Decoder design.} Different from image classification, semantic segmentation requires to output a high-resolution segmentation mask. We ablate three
different decoder structures, which are shown in \cref{fig:decoder designs}, corresponding to the architectures of SegFormer, SETR, our IncepFormer, respectively. 
The results are listed in \cref{fig:decoder designs}, where we can see that IncepFormer-T(c) achieves the best performance and the computational cost is also low. In particular,
IncepFormer-T(c) yields 0.2 mIoU improvement compared to IncepFormer-T(a). We speculate that the first MLP-layer in SegFormer decoder, where features from the encoder are unified to
the same channel dimension, may lose the channel information.

\subsection{Comparison with state-of-the-art methods}
In this section, we divide the existing approaches into two classes, where one is the transformer-base methods, and the other is the CNN-based models. We compare our IncepFormer with
transformer-based methods on ADE20K, Cityscapes and COCO-Stuff, and CNN-based models on ADE20K, Cityscapes, Pascal Context and Pascal VOC, respectively.

\noindent\textbf{Comparison with state-of-the-art transformer models.} We compare IncepFormer with state-of-the-art transformer-based semantic segmentation methods, such as
SegFormer, HRFormer, MaskFormer, Mask2Former and SETR. As illustrated in \cref{table:transformer-based comparison}, our IncepFormer-T yields similar results compared to HRFormer 
but only introducing 19\% computational cost on the ADE20K dataset. Moreover, our IncepFormer-S outperforms SegFormer-B2 (81.6\% vs. 81.0\%) while using only 1/8 (83.2G vs. 717.1G) computations when dealing with high-resolution urban scenes from the Cityscapes dataset. We also show a qualitative comparison with SegFormer in \cref{figure:compare figure}.

\noindent\textbf{Comparison with state-of-the-art CNN models.} As shown in \cref{table:cnns-based methods} and \cref{table:Pascal Context}, we compare our IncepFormer with state-of-the-art 
CNN-based segmentation models, like FCN, EncNetm, DeeplabV3 and SegNext. IncepFormer-B outperforms the popular HRNet (OCR) model (57.78\% vs. 56.2\%) with even less parameters and computations,
which is designed for segmentation task elaborately on the Pascal Context dataset. Due to the unavailability of the pretrained model, the result for SegNext-S on Pascal VOC dataset is obtained with the same setting as~\cite{guo2022segnext} but running on our GPUs.

\noindent\textbf{Balance in performance and model parameter.} ADE20K is a challenging and widely used dataset in semantic segmentation. As shown in \cref{fig:F1}, we plot the 
performance-parameter curves of different methods on ADE20K validation set. Clearly, our method obtains the best balance between performance and computational cost, compared with some
famous state-of-the-art methods, such as Swin Transformer, SETR and SegFormer.

\section{Conclusion}
In this paper, we present IncepFormer, a simple, efficient and powerful semantic segmentation method which contains a pyramid transformer encoder with Inception self-attention and an effective FFN, and a lightweight Upsample-Concat decoder. The proposed Incep-MHSA  pays attention to the use of multi-scale strip-shaped convolution, which yields  better local  feature extraction and spatial reduction attention modeling, while having a huge complexity saving. Experiment results demonstrate that IncepFormer surpasses current state-of-the-art transformer-based and CNN-based methods by a considerable
margin. We hope our method can serve as baseline and provides inspiration for further research in semantic segmentation. 
Future work may include how to extend the IncepFormer to large-scale model with probably 100M+ parameters and how to transfer it to other vision tasks.

{\small
\bibliographystyle{ieee_fullname}
\bibliography{egbib}
}

\clearpage
\onecolumn
\begin{@twocolumnfalse}
\section*{\centering{ \Large Supplementary Material for the paper ``IncepFormer: Efficient Inception Transformer with Pyramid Pooling for Semantic Segmentation"} }
\vspace{25mm}

In this document, we firstly provide the specification for the proposed IncepFormer. We have designed three versions for IncepFormer across model scale, i.e., tiny version (IPT-T), small version (IPT-S), and basic version (IPT-B). The detailed settings of these three versions are presented in \cref{table:encoder setting}, where they differ from each other in the number of inception transformer blocks used at each stage and also the output channel count in the decoder.

Secondly, we validate the performance of these three variants on four datasets. The results are tabulated in \cref{tabel:ablation 1}. As can be observed, our decoder is sufficiently lightweight, and the parameter size of all decoders  are less than 1M. In addition, consistent improvements are yielded on all datasets with the increase of the encoder size. Our largest model IPT-B achieves mIoUs of 49.4\%/50.2\%, 82.0\%/82.9\%, 46.4\%/46.7\%, 56.9\%/57.8\%, on ADE20K, Cityscape, COCO-Stuff, Pascal Context, respectively, under the SS/MS configuration. Compared to other two smaller models, IPT-B generally has the heaviest computations, which induces the largest computational complexity on the Cityscapes dataset due to the large spatial resolution of the images. 

Thirdly, \Cref{fig1} and \ref{fig2} present the qualitative comparison between our IncepFormer-S and SegFormer-B2 on Cityscapes and ADE20K dataset, respectively. We can see our proposed IncepFormer achieves better segmentation results. In \cref{fig1}, our proposed model can better segment the strip-shaped objects, such as road light, line pole, etc, due to the design of inception transformer block. In \cref{fig2}, for the more challenging dataset ADE20K,  we observe that IncepFormer-S provides more consistent lables in large instances and handle partial occlusion well. 

\begin{table*}[htbp]
	\centering
	\renewcommand{\arraystretch}{1.7}
	\begin{tabular}{c|p{2.6cm}<{\centering}|p{0.5cm}<{\centering}|p{2.5cm}<{\centering}|p{2.5cm}<{\centering}|p{2.5cm}<{\centering}}
		\toprule[2pt]
		Stage & Output Size & R & IPT-T & IPT-S & IPT-B \\
		\hline
		
		1 & $\frac{H}{4} \times \frac{W}{4} \times C_1$ & 8 & $C_1$ = 64, \,$D$=2 & $C_1$ = 64, \,$D$=3 & $C_1$ = 64, \,$D$=3 \\ 
		\hline
		2 & $\frac{H}{8} \times \frac{W}{8} \times C_2$ & 4 & $C_2$ = 128, \,$D$=2 & $C_2$ = 128, \,$D$=4 & $C_2$ = 128, \,$D$=6 \\ 
		\hline
		3 & $\frac{H}{16} \times \frac{W}{16} \times C_3$ & 2 & $C_3$ = 320, \,$D$=4 & $C_3$ = 320, \,$D$=12 & $C_3$ = 320, \,$D$=24 \\ 
		\hline
		4 & $\frac{H}{32} \times \frac{W}{32} \times C_4$ & 1 & $C_4$ = 512, \,$D$=2 & $C_4$ = 512, \,$D$=3 & $C_4$ = 512, \,$D$=2 \\ 
		\hline
		\multicolumn{3}{c|}{\small Output channels of $\rm{Conv}^{1}$ in decoder} & 512 & 768 & 768\\
		\hline
		\multicolumn{3}{c|}{Model Parameters (M)} & 14.0  & 24.6 & 39.6 \\
		\bottomrule[2pt]
	\end{tabular}
	\caption{Detailed settings of the three versions of our proposed Inception Transformer. In this table, ``R'' is the reduction ratio in the Incep-MHSA. 
		``$C_i$'' and ``$D$'' denote the numbers of channels and Inception transformer blocks at each stage, respectively. ``Output channels of ${\rm Conv}^{1}$'' represents 
		the output channels of the convolution with the kernel size of $1 \times 1$ in decoder. At last, ``Parameters'' are calculated 
		on the ADE20K dataset. Note that, the scale of model parameters may vary slightly due to different dataset having various number of categories.}
	\label{table:encoder setting}
\end{table*}

\begin{table*}[htbp]
	\centering
	\renewcommand{\arraystretch}{1.85}
	\resizebox{\textwidth}{!}{
		\begin{tabular}{c|c|c|c|c|c|c|c|c|c|c|c|c|c|c}
			\toprule[1pt]
			Encoder & \multicolumn{2}{c|}{Params(M)} & \multicolumn{3}{c|}{ADE20K} & \multicolumn{3}{c|}{Cityscapes} & \multicolumn{3}{c|}{COCO-Stuff} & \multicolumn{3}{c}{Pascal Context} \\
			\cline{2-15}
			Model Size & Encoder & Decoder & GFLOPs & \multicolumn{2}{c|}{mIoU(SS/MS)} & GFLOPs & \multicolumn{2}{c|}{mIoU(SS/MS)} & GFLOPs & \multicolumn{2}{c|}{mIoU(SS/MS)} & GFLOPs & \multicolumn{2}{c}{mIoU(SS/MS)}\\
			\midrule
			IPT-T & 13.5 & 0.5 & 21.2 & 44.8 & 46.3 & 45.4 & 80.5 & 81.8 & 21.4 & 43.8 & 44.0 & 18.0 & 53.8 & 55.1 \\
			\midrule
			IPT-S & 23.8 & 0.8 & 38.5 & 47.7 & 49.0 & 83.2 & 81.6 & 82.6 & 38.7 & 45.4 & - & 32.9 & 55.3 & 56.4 \\
			\midrule
			IPT-B & 38.8 & 0.8 & 54.6 & 49.4 & 50.2 & 119.3 & 82.0 & 82.9 & 54.8 & 46.4 & 46.7 & 47 & 56.9 & 57.8 \\
			
			\bottomrule[1pt]
	\end{tabular}}
	\caption{Performance on our IncepFormer with three different sizes. The number of parameters(M) and FLOPs(G) is calculated on the input size of $512 \times 512$ for ADE20K and COCO-Stuff, 
		$768 \times 768$ for Cityscapes and $480 \times 480$ for Pascal Context. ``SS'' and ``MS'' mean single/multi-scale test.}
	\label{tabel:ablation 1}
\end{table*}

 \begin{figure*}[htbp]
	\centering
	\includegraphics[width=1.0\linewidth]{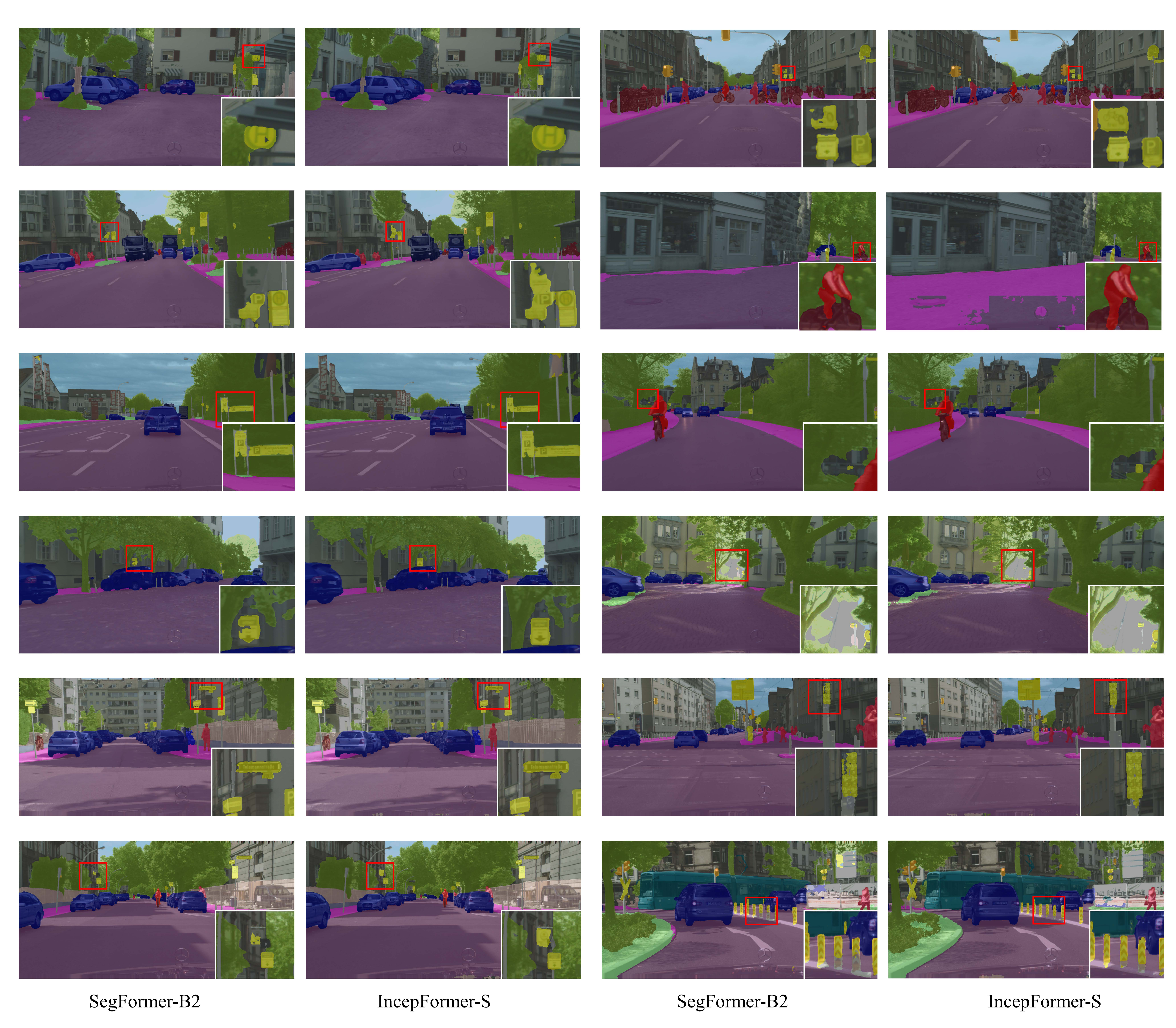}
	\caption{Qualitative comparison of SegFormer-B2 and IncepFormer-S on the Cityscapes dataset}
	\label{fig1}
\end{figure*}

\begin{figure*}[htbp]
	\centering
	\includegraphics[width=0.97\linewidth]{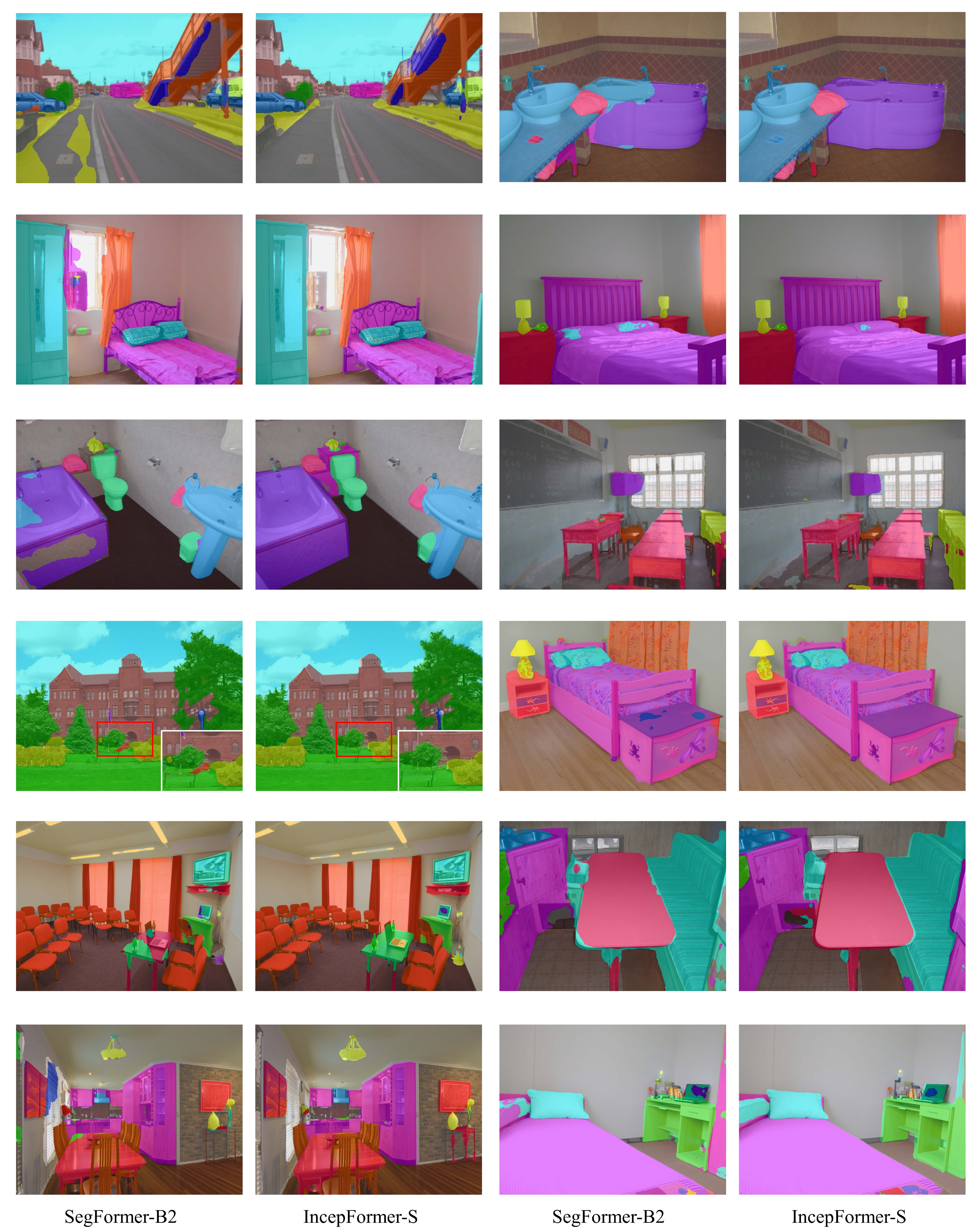}
	\caption{Qualitative comparison of SegFormer-B2 and IncepFormer-S on the ADE20K dataset}
	\label{fig2}
\end{figure*}

\end{@twocolumnfalse}

\end{document}